\documentclass[11pt,a4paper]{article}
\PassOptionsToPackage{breaklinks}{hyperref}
\usepackage[hyperref]{emnlp2020}
\usepackage{times}
\usepackage{multirow}
\usepackage{dirtytalk}
\usepackage{graphicx}
\usepackage{latexsym}

\usepackage{microtype}

\aclfinalcopy % Uncomment this line for the final submission
\title{iNLTK: Natural Language Toolkit for Indic Languages}

\author{Gaurav Arora \\
        Jio Haptik \\
  \texttt{gaurav@haptik.ai}}

\date{}

\begin{document}
\maketitle
\begin{abstract}
We present iNLTK, an open-source NLP library consisting of pre-trained language models and out-of-the-box support for Data Augmentation, Textual Similarity, Sentence Embeddings, Word Embeddings, Tokenization and Text Generation in 13 Indic Languages. By using pre-trained models from iNLTK for text classification on publicly available datasets, we significantly outperform previously reported results. On these datasets, we also show that by using pre-trained models and data augmentation from iNLTK, we can achieve more than 95\% of the previous best performance by using less than 10\% of the training data. iNLTK is already being widely used by the community and has 40,000+ downloads, 600+ stars and 100+ forks on GitHub. The library is available at https://github.com/goru001/inltk.
\end{abstract}

\section{Introduction}

Deep learning offers a way to harness large amounts of computation and data with little engineering by hand \citep{article}. With distributed representation, various deep models have become the new state-of-the-art methods for NLP problems. Pre-trained language models \citep{devlin-etal-2019-bert} can model syntactic/semantic relations between words and reduce feature engineering. These pre-trained models are useful for initialization and/or transfer learning for NLP tasks. Pre-trained models are typically learned using unsupervised approaches from large, diverse monolingual corpora \citep{kunchukuttan2020indicnlpcorpus}. While we have seen exciting progress across many tasks in natural language processing over the last years, most such results have been achieved in English and a small set of other high-resource languages \citep{ruder2020}.

Indic languages, widely spoken by more than a billion speakers, lack pre-trained deep language models, trained on a large corpus, which can provide a headstart for downstream tasks using transfer learning. Availability of such models is critical to build a system that can achieve good results in \say{low-resource} settings - where labeled data is scarce and computation is expensive, which is the biggest challenge for working on NLP in Indic Languages. Additionally, there's lack of Indic languages support in NLP libraries like spacy\footnote{https://spacy.io/}, nltk\footnote{https://www.nltk.org/} - creating a barrier to entry for working with Indic languages.

iNLTK, an open-source natural language toolkit for Indic languages, is designed to address these problems and to significantly lower barriers to doing NLP in Indic Languages by

\begin{itemize}

\item sharing pre-trained deep language models, which can then be fine-tuned and used for downstream tasks like text classification,

\item providing out-of-the-box support for Data Augmentation, Textual Similarity, Sentence Embeddings, Word Embeddings, Tokenization and Text Generation built on top of pre-trained language models, lowering the barrier for doing applied research and building products in Indic languages
\end{itemize}

iNLTK library supports 13 Indic languages, including English, as shown in Table \ref{tab:supported-languages}. GitHub repository\footnote{https://github.com/goru001/inltk} for the library contains source code, links to download pre-trained models, datasets and API documentation\footnote{https://inltk.readthedocs.io/}. It includes reference implementations for reproducing text-classification results shown in Section \ref{sec:text-classification-eval}, which can also be easily adapted to new data. The library has a permissive MIT License and is easy to download and install via pip or by cloning the GitHub repository. 

% Table for statistics of wiki dataset

\begin{table*}[]
\centering
\begin{tabular}{ccccc}
\hline
\textbf{Language} & \multicolumn{2}{c}{\textbf{\# Wikipedia Articles}} & \multicolumn{2}{c}{\textbf{\# Tokens}} \\
\textbf{} & \textbf{Train} & \textbf{Valid} & \textbf{Train} & \textbf{Valid} \\ \hline
Hindi & 137,823 & 34,456 & 43,434,685 & 10,930,403 \\
Bengali & 50,661 & 21,713 & 15,389,227 & 6,493,291 \\
Gujarati & 22,339 & 9,574 & 4,801,796 & 2,005,729 \\
Malayalam & 8,671 & 3,717 & 1,954,174 & 926,215 \\
Marathi & 59,875 & 25,662 & 7,777,419 & 3,302,837 \\
Tamil & 102,126 & 25,255 & 14,923,513 & 3,715,380 \\
Punjabi & 35,637 & 8,910 & 9,214,502 & 2,276,354 \\
Kannada & 26,397 & 6,600 & 11,450,264 & 3,110,983 \\
Oriya & 12,446 & 5,335 & 2,391,168 & 1,082,410 \\
Sanskrit & 18,812 & 6,682 & 11,683,360 & 4,274,479 \\
Nepali & 27,129 & 11,628 & 3,569,063 & 1,560,677 \\
Urdu & 107,669 & 46,145 & 15,421,652 & 6,773,909 \\ \hline
\end{tabular}
\caption{Statistics of Wikipedia Articles Dataset used for training Language Models}
\label{tab:wiki_statistics}
\end{table*}

% Table for supported languages

\begin{table}[h]
\centering
\begin{tabular}{ll|ll}
\hline
\textbf{Language} & \textbf{Code} & \textbf{Language} & \textbf{Code} \\ \hline
Hindi & hi & Marathi & mr \\
Punjabi & pa & Bengali & bn \\
Gujarati & gu & Tamil & ta \\
Kannada & kn & Urdu & ur \\
Malayalam & ml & Nepali & ne \\
Oriya & or & Sanskrit & sa \\
English & en &  &  \\ \hline
\end{tabular}
\caption{Languages supported in iNLTK}
\label{tab:supported-languages}
\end{table}

\section{iNLTK Pretrained Language Models}

iNLTK has pre-trained ULMFiT \citep{DBLP:journals/corr/abs-1801-06146} and TransformerXL \citep{DBLP:journals/corr/abs-1901-02860} language models for 13 Indic languages. All the language models (LMs) were trained from scratch using PyTorch \citep{Paszke2017AutomaticDI} and Fastai\footnote{https://github.com/fastai/fastai}, except for English. Pre-trained LMs were then evaluated on downstream task of text classification on public datasets. Pre-trained LMs for English were borrowed from Fastai directly. This section describes training of language models and their evaluation.

\subsection{Dataset preparation}

We obtained a monolingual corpora for each one of the languages from Wikipedia for training LMs from scratch. We used the wiki extractor\footnote{https://github.com/attardi/wikiextractor} tool and BeautifulSoup\footnote{https://www.crummy.com/software/BeautifulSoup} for text extraction from Wikipedia. Wikipedia articles were then cleaned and split into train-validation sets. Table \ref{tab:wiki_statistics} shows statistics of the monolingual Wikipedia articles dataset for each language. Hindi Wikipedia articles dataset is the largest one, while Malayalam and Oriya Wikipedia articles datasets have the least number of articles.

% vocab size table

\begin{table}[]
\centering
\begin{tabular}{cc|cc}
\hline
\textbf{Language} & \textbf{\begin{tabular}[c]{@{}c@{}}Vocab\\  size\end{tabular}} & \textbf{Language} & \textbf{\begin{tabular}[c]{@{}c@{}}Vocab\\  size\end{tabular}} \\ \hline
Hindi & 30,000 & Marathi & 30,000 \\
Punjabi & 30,000 & Bengali & 30,000 \\
Gujarati & 20,000 & Tamil & 8,000 \\
Kannada & 25,000 & Urdu & 30,000 \\
Malayalam & 10,000 & Nepali & 15,000 \\
Oriya & 15,000 & Sanskrit & 20,000 \\ \hline
\end{tabular}
\caption{Vocab size for languages supported in iNLTK}
\label{tab:vocab-size}
\end{table}

% Text classification accuracy table

\begin{table*}[]
\centering
\begin{tabular}{cccccc}
\hline
\textbf{Language} & \textbf{Dataset} & \textbf{FT-W} & \textbf{FT-WC} & \textbf{INLP} & \textbf{iNLTK} \\ \hline
\multirow{3}{*}{Hindi} & BBC Articles & 72.29 & 67.44 & 74.25 & \textbf{78.75} \\
 & IITP+Movie & 41.61 & 44.52 & 45.81 & \textbf{57.74} \\
 & IITP Product & 58.32 & 57.17 & 63.48 & \textbf{75.71} \\ \hline
Bengali & Soham Articles & 62.79 & 64.78 & 72.50 & \textbf{90.71} \\ \hline
Gujarati & \multirow{4}{*}{\begin{tabular}[c]{@{}c@{}}iNLTK \\ Headlines\end{tabular}} & 81.94 & 84.07 & 90.90 & \textbf{91.05} \\
Malayalam &  & 86.35 & 83.65 & 93.49 & \textbf{95.56} \\
Marathi &  & 83.06 & 81.65 & 89.92 & \textbf{92.40} \\
Tamil &  & 90.88 & 89.09 & 93.57 & \textbf{95.22} \\ \hline
Punjabi & \multirow{3}{*}{\begin{tabular}[c]{@{}c@{}}IndicNLP News\\  Category \\  \end{tabular}} & 94.23 & 94.87 & 96.79 & \textbf{97.12} \\
Kannada &  & 96.13 & 96.50 & 97.20 & \textbf{98.87} \\
Oriya &  & 94.00 & 95.93 & 98.07 & \textbf{98.83} \\ \hline
\end{tabular}
\caption{Text classification accuracy on public datasets}
\label{tab:classification-results}
\end{table*}

% Table for LM Perplexities

\begin{table}[]
\centering
\begin{tabular}{ccc}
\hline
\textbf{Language} & \multicolumn{2}{c}{\textbf{Perplexity}} \\
\textbf{} & \textbf{ULMFiT} & \textbf{TransformerXL} \\ \hline
Hindi & 34.0 & 26.0 \\
Bengali & 41.2 & 39.3 \\
Gujarati & 34.1 & 28.1 \\
Malayalam & 26.3 & 25.7 \\
Marathi & 17.9 & 17.4 \\
Tamil & 19.8 & 17.2 \\
Punjabi & 24.4 & 14.0 \\
Kannada & 70.1 & 61.9 \\
Oriya & 26.5 & 26.8 \\
Sanskrit & 5.5 & 2.7 \\
Nepali & 31.5 & 29.3 \\
Urdu & 13.1 & 12.5 \\ \hline
\end{tabular}
\caption{Perplexity on validation set of Language Models in iNLTK }
\label{tab:perplexity}
\end{table}

\subsection{Tokenization}
\label{subsec:tokenization}

We create subword vocabulary for each one of the languages by training a SentencePiece\footnote{https://github.com/google/sentencepiece} tokenization model on Wikipedia articles dataset, using unigram segmentation algorithm \citep{DBLP:journals/corr/abs-1808-06226}. An important property of SentencePiece tokenization, necessary for us to obtain a valid subword-based language model, is its reversibility. We do not use subword regularization as the available training dataset is large enough to avoid overfitting. Table \ref{tab:vocab-size} shows subword vocabulary size of the tokenization model for each one of the languages.

\subsection{Language Model Training}

Our model is based on the Fastai implementation of ULMFiT and TransformerXL. Hyperparameters of the final model are accessible from the GitHub repository of the library. Table \ref{tab:perplexity} shows perplexity of language models on validation set. TransformerXL consistently performs better for all languages.

\subsection{Text Classification Evaluation}
\label{sec:text-classification-eval}

We evaluated pre-trained ULMFiT language models on downstream task of text-classification using following publicly available datasets: (a) IIT-Patna Sentiment Analysis dataset \citep{akhtar-etal-2016-hybrid}, (b) BBC News Articles classification dataset\footnote{https://github.com/NirantK/hindi2vec/releases/tag/bbc-hindi-v0.1}, (c) iNLTK Headlines dataset\footnote{https://github.com/goru001/inltk}, (d) Soham Bengali News classification dataset\footnote{https://www.kaggle.com/csoham/classification-bengali-news-articles-indicnlp}, (e) IndicNLP News Category classification dataset \citep{kunchukuttan2020indicnlpcorpus}. Train and test splits, derived by the authors \citep{kunchukuttan2020indicnlpcorpus} from the above mentioned corpora and used for benchmarking, are available on the IndicNLP corpus website\footnote{https://github.com/AI4Bharat/indicnlp\_corpus}. Table \ref{tab:stats-pub-class} shows statistics of these datasets.

% Statistics of classification datasets

\begin{table}[]
\centering
\resizebox{\columnwidth}{!}{%
\begin{tabular}{ccccc}
\hline
\multirow{2}{*}{\textbf{Language}} & \multirow{2}{*}{\textbf{Dataset}} & \multirow{2}{*}{\textbf{N}} & \multicolumn{2}{c}{\textbf{\# Examples}} \\ \cline{4-5} 
 &  &  & \textbf{Train} & \textbf{Test} \\ \hline
Hindi & BBC Articles & 6 & 3467 & 866 \\
 & IITP+Movie & 3 & 2480 & 310 \\
 & IITP Product & 3 & 4182 & 523 \\ \hline
Bengali & \begin{tabular}[c]{@{}c@{}}Soham \\ Articles\end{tabular} & 6 & 11284 & 1411 \\ \hline
Gujarati & \multirow{4}{*}{\begin{tabular}[c]{@{}c@{}}iNLTK\\ Headlines\end{tabular}} & 3 & 5269 & 659 \\
Malayalam &  & 3 & 5036 & 630 \\
Marathi &  & 3 & 9672 & 1210 \\
Tamil &  & 3 & 5346 & 669 \\ \hline
Punjabi & \multirow{3}{*}{\begin{tabular}[c]{@{}c@{}}IndicNLP\\  News\\ Category\end{tabular}} & 4 & 2496 & 312 \\
Kannada &  & 3 & 24000 & 3000 \\
Oriya &  & 4 & 24000 & 3000 \\ \hline
\end{tabular}%
}
\caption{Statistics of publicly available classification datasets  (N is the number of classes)}
\label{tab:stats-pub-class}
\end{table}

% Data Augmentation Table

\begin{table*}[h]
\centering
\resizebox{\textwidth}{!}{%
\begin{tabular}{ccccccccc}
\hline
\textbf{Language} & \textbf{Dataset} & \multicolumn{2}{c}{\textbf{\begin{tabular}[c]{@{}c@{}}\# Training\\  Examples\end{tabular}}} & \textbf{\begin{tabular}[c]{@{}c@{}}\%age\\ reduction\end{tabular}} & \textbf{\begin{tabular}[c]{@{}c@{}}INLP\\ Accuracy\end{tabular}} & \multicolumn{3}{c}{\textbf{iNLTK Accuracy}} \\ \cline{3-4} \cline{6-9} 
\textbf{} & \textbf{} & \textbf{Full} & \textbf{Reduced} & \textbf{} & \textbf{Full} & \textbf{Full} & \multicolumn{2}{c}{\textbf{Reduced}} \\ \cline{8-9} 
\textbf{} & \textbf{} & \textbf{} & \textbf{} & \textbf{} & \textbf{} & \textbf{} & \textbf{\begin{tabular}[c]{@{}c@{}}Without\\ Data Aug\end{tabular}} & \textbf{\begin{tabular}[c]{@{}c@{}}With\\ Data Aug\end{tabular}} \\ \hline
Hindi & IITP+Movie & 2,480 & 496 & 80\% & 45.81 & 57.74 & 47.74 & 56.13 \\ \hline
Bengali & \begin{tabular}[c]{@{}c@{}}Soham\\ Articles\end{tabular} & 11,284 & 112 & 99\% & 72.50 & 90.71 & 69.88 & 74.06 \\ \hline
Gujarati & \multirow{4}{*}{\begin{tabular}[c]{@{}c@{}}iNLTK\\ Headlines\end{tabular}} & 5,269 & 526 & 90\% & 90.90 & 91.05 & 80.88 & 81.03 \\
Malayalam &  & 5,036 & 503 & 90\% & 93.49 & 95.56 & 82.38 & 84.29 \\
Marathi &  & 9,672 & 483 & 95\% & 89.92 & 92.40 & 84.13 & 84.55 \\
Tamil &  & 5,346 & 267 & 95\% & 93.57 & 95.22 & 86.25 & 89.84 \\ \hline
 & Average & 6514.5 & 397.8 & 91.5\% & 81.03 & 87.11 & 75.21 & 78.31 \\ \hline
\end{tabular}%
}
\caption{Comparison of Accuracy on INLP trained on Full Training set vs Accuracy on iNLTK, using data augmentation, trained on reduced training set }
\label{tab:data-aug-results}
\end{table*}

iNLTK results were compared against results reported in \citep{kunchukuttan2020indicnlpcorpus} for pre-trained embeddings released by the FastText project trained on Wikipedia (FT-W) \citep{DBLP:journals/corr/BojanowskiGJM16}, Wiki+CommonCrawl (FT-WC) \citep{DBLP:journals/corr/abs-1802-06893} and INLP embeddings \citep{kunchukuttan2020indicnlpcorpus}. Table \ref{tab:classification-results} shows that iNLTK significantly outperforms other models across all languages and datasets\footnote{\label{reproducibility-footnote}Refer GitHub repository of the library for instructions to reproduce results}.

\section{iNLTK API}

iNLTK is designed to be simple for practitioners in order to lower the barrier for doing applied research and building products in Indic languages. This section discusses various NLP tasks for which iNLTK provides out-of-the-box support, under a unified API.

\textbf{Data Augmentation} helps in improving the performance of NLP models \citep{duboue2006answering,marton2009improved}. It is even more important in \say{low-resource} settings, where labeled data is scarce. iNLTK provides augmentations\footnote{https://inltk.readthedocs.io/en/latest/api\_docs.html\#get-similar-sentences} for a sentence while preserving its semantics following a two step process. Firstly, it generates candidate paraphrases by replacing original sentence tokens with tokens which have closest embeddings from the embedding layer of pre-trained language model. And then, it chooses top paraphrases which are similar to original sentence, where similarity between sentences is calculated as the cosine similarity of sentence embeddings, obtained from pre-trained language model's encoder.

To evaluate the effectiveness of using data augmentation from iNLTK in low resource settings, we prepare\footnote{Notebooks to prepare reduced datasets are accessible from the GitHub repository of the library} \textbf{reduced train sets} of publicly available text-classification datasets by picking first $N$ examples from the full train set\footnote{Labels in publicly available full train sets were not grouped together, instead were randomly shuffled}, where $N$ is equal to size of reduced train set and compare accuracy of the classifier trained \textit{with} vs \textit{without} data augmentation. Table \ref{tab:data-aug-results} shows reduced dataset statistics and comparison of results obtained on full and reduced datasets using iNLTK. Using data augmentation from iNLTK gives significant increase in accuracy on Hindi, Bengali, Malayalam and Tamil dataset, and minor improvements in Gujarati and Marathi datasets. Additionally, Table \ref{tab:data-aug-results} compares previously obtained best results on these datasets using INLP embeddings \citep{kunchukuttan2020indicnlpcorpus} with results obtained using iNLTK pretrained models and iNLTK's data augmentation utility. On an average, with iNLTK we are able to achieve more than 95\% of the previous accuracy using less than 10\% of the training data\footnote{Refer GitHub repository of the library for instructions to reproduce results on full and reduced dataset}.

\textbf{Semantic Textual Similarity} (STS) assesses the degree to which the underlying semantics of two segments of text are equivalent to each other \citep{agirre-etal-2016-semeval}. iNLTK compares\footnote{https://inltk.readthedocs.io/en/latest/api\_docs.html\#get-sentence-similarity} sentence embeddings of the two segments of text, obtained from pre-trained language model's encoder, using a comparison function, to evaluate semantic textual similarity. Cosine similarity between sentence embeddings is used as the default comparison function.

\textbf{Distributed representations} are the cornerstone of modern NLP, which have led to significant advances in many NLP tasks. iNLTK provides utilities to obtain distributed representations for \textbf{words}\footnote{https://inltk.readthedocs.io/en/latest/api\_docs.html\#get-embedding-vectors}, \textbf{sentences and documents}\footnote{https://inltk.readthedocs.io/en/latest/api\_docs.html\#get-sentence-encoding} obtained from embedding layer and encoder output of pre-trained language models, respectively.

Additionally, iNLTK provides utilities to \textbf{generate text}\footnote{https://inltk.readthedocs.io/en/latest/api\_docs.html\#predict-next-n-words} given a prompt, using pre-trained language models, \textbf{tokenize}\footnote{https://inltk.readthedocs.io/en/latest/api\_docs.html\#tokenize} text using sentencepiece tokenization models described in Section \ref{subsec:tokenization},  identify\footnote{https://inltk.readthedocs.io/en/latest/api\_docs.html\#identify-language} which one of the supported Indic languages is given \texttt{text} in and remove tokens of a foreign language\footnote{https://inltk.readthedocs.io/en/latest/api\_docs.html\#remove-foreign-languages} from given \texttt{text}. 

\section{Related Work}

NLP and ML communities have a strong culture of building open-source tools. There are lots of easy-to-use, user-facing libraries for general-purpose NLP like NLTK \citep{loper}, Stanford CoreNLP \citep{manning-etal-2014-stanford}, Spacy \citep{spacy2}, AllenNLP \citep{DBLP:journals/corr/abs-1803-07640}, Flair \citep{akbik-etal-2019-flair}, Stanza \citep{qi2020stanza} and Huggingface Transformers \citep{Wolf2019HuggingFacesTS}. But most of these libraries have limited or no support for Indic languages, creating a barrier to entry for working with Indic languages. Additionally, for many Indic languages word embeddings have been trained, but they still lack richer pre-trained representations from deep language models \citep{kunchukuttan2020indicnlpcorpus}. iNLTK tries to solve these problems by providing pre-trained language models and out-of-the-box support for a variety of NLP tasks in 13 Indic languages.

\section{Conclusion and Future Work}

iNLTK provides pre-trained language models and supports Data Augmentation, Textual Similarity, Sentence Embeddings, Word Embeddings, Tokenization and Text Generation in 13 Indic Languages. Our results significantly outperform other methods on text-classification benchmarks, using pre-trained models from iNLTK. These pre-trained models from iNLTK can be used as-is for a variety of NLP tasks, or can be fine-tuned on domain specific datasets. iNLTK is being widely\footnote{https://github.com/goru001/inltk/network/members} used\footnote{https://pepy.tech/project/inltk} and appreciated\footnote{https://github.com/goru001/inltk/stargazers} by the community\footnote{https://github.com/goru001/inltk\#inltks-appreciation}.

We are working on expanding the supported languages in iNLTK to include other Indic languages like Telugu, Maithili; code mixed languages like Hinglish (Hindi and English), Manglish (Malayalam and English) and Tanglish (Tamil and English); expanding supported model architectures to include BERT. Additionally, we want to mitigate any possible unwarranted biases which might exist in pre-trained language models \citep{lu2019gender}, because of training data, which might propagate into downstream systems using these models. While these tasks are work in progress, we hope this library will accelerate NLP research and development in Indic languages.

\section*{Acknowledgments}

We are thankful to Anurag Singh\footnote{https://github.com/anuragshas} and Ravi Annaswamy\footnote{https://github.com/ravi-annaswamy} for their contributions to support Urdu and Tamil in the iNLTK library, respectively.

\bibliographystyle{acl_natbib}
\bibliography{emnlp2020}

\end{document}